\documentclass[acmtog,authorversion,nonacm,screen]{acmart}

\usepackage{graphicx}
\usepackage{booktabs} 
\usepackage[ruled]{algorithm2e} 
\usepackage{multirow}

\SetAlFnt{\small}
\SetAlCapFnt{\small}
\SetAlCapNameFnt{\small}
\SetAlCapHSkip{0pt}

\makeatletter
\let\@authorsaddresses\@empty
\makeatother

\makeatletter
\def\@ACM@checkaffil{
    \if@ACM@instpresent\else
    \ClassWarningNoLine{\@classname}{No institution present for an affiliation}%
    \fi
    \if@ACM@citypresent\else
    \ClassWarningNoLine{\@classname}{No city present for an affiliation}%
    \fi
    \if@ACM@countrypresent\else
        \ClassWarningNoLine{\@classname}{No country present for an affiliation}%
    \fi
}
\makeatother

\begin{document}
\title{Interactive Segment Anything NeRF with Feature Imitation}

\author{Xiaokang Chen}
\affiliation{
 \institution{School of Intelligence Science and Technology, Peking University}
}

\author{Jiaxiang Tang}
\affiliation{
 \institution{School of Intelligence Science and Technology, Peking University}
}

\author{Diwen Wan}
\affiliation{
 \institution{School of Intelligence Science and Technology, Peking University}
}

\author{Jingbo Wang}
\affiliation{
 \institution{The Chinese University of Hong Kong}
}

\author{Gang Zeng}
\affiliation{
 \institution{School of Intelligence Science and Technology, Peking University}
}

\begin{abstract}
This paper investigates the potential of enhancing Neural Radiance Fields (NeRF) with semantics to expand their applications. Although NeRF has been proven useful in real-world applications like VR and digital creation, the lack of semantics hinders interaction with objects in complex scenes. We propose to imitate the backbone feature of off-the-shelf perception models to achieve zero-shot semantic segmentation with NeRF. Our framework reformulates the segmentation process by directly rendering semantic features and only applying the decoder from perception models. This eliminates the need for expensive backbones and benefits 3D consistency. Furthermore, we can project the learned semantics onto extracted mesh surfaces for real-time interaction. With the state-of-the-art Segment Anything Model (SAM), our framework accelerates segmentation by 16 times with comparable mask quality. The experimental results demonstrate the efficacy and computational advantages of our approach. Project page: \url{https://me.kiui.moe/san/}.
\end{abstract}

\keywords{NeRF, Interactive 3D Segmentation}

\begin{teaserfigure}
    \centering
    \includegraphics[width=1.00 \linewidth]{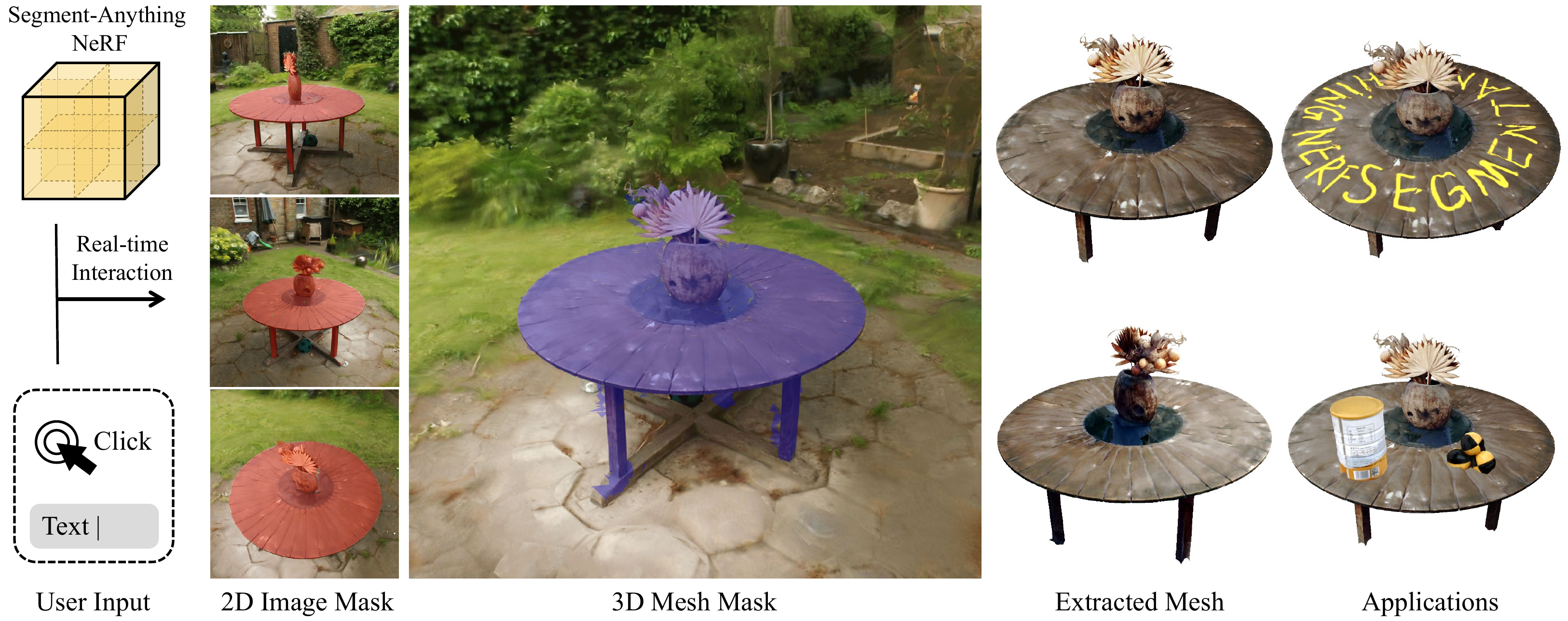}
    \caption{Our pipeline allows click- or text-based user interaction to perform zero-shot semantic segmentation in 3D space. We further investigate single-object mesh extraction by projecting semantic masks onto mesh surface, leading to applications like texture editing and model composition.}
    \label{fig:teaser}
    \vspace{0.5cm}
\end{teaserfigure}

\maketitle

\footnotetext[1]{X. Chen and J. Tang have made equal contributions in the technical aspect. D. Wan and J. Wang have made equal contributions in the writing. The authorship order is determined based on the alphabetical order of the authors' names.}

\section{Introduction}

Neural Radiance Fields (NeRF)~\cite{mildenhall2020nerf} has recently garnered significant attention as a promising method for synthesizing photo-realistic images of complex 3D environments. 
By providing a theoretically sound approach to scene reconstruction from 2D images, NeRF demonstrates the potential to bridge the gap between captured images and the corresponding 3D world with more comprehensive information. 
Despite their strengths, the output of NeRFs remains limited to geometry and appearance, devoid of any explicit semantic information. This lack of interpretability can hamper the development of flexible applications aimed at interacting with  content in reconstructed scenarios, such as editing the appearance of specific objects in a complex 3D environment or extracting the mesh of this object.

In this paper, we address the challenge of incorporating explicit semantic information into the otherwise purely geometric representation of reconstructed scenes provided by Neural Radiance Fields. 
The goal is to enhance the flexibility and potential of NeRF for interactive applications by leveraging off-the-shelf semantic models, such as pre-trained large-scale perception~\cite{SAM,X-Decoder} and language models~\cite{CLIP}, to infuse semantically meaningful information into their output. 
To this end, we propose to use segmentation mask-based approaches that enable pixel-wise classification of objects in the rendered images of the reconstructed scene, and thus allow for object-level human scene interactions. 
As shown in Figure~\ref{fig:teaser}, our method demonstrates the effectiveness of using pre-existing semantic knowledge to enhance the output of NeRF with semantics, thereby improving their applicability in complex real-world scenarios.

Our proposed framework is built upon the grid-based NeRF~\cite{mueller2022instant,sun2021direct,tang2022nerf2mesh} representation, which has gained significant popularity across various fields~\cite{ISRF,Magic3D,lerf2023}.
Although one plausible method for NeRF to cooperate with off-the-shelf perception and large language models is to extract semantics directly from the rendered images, this approach can often be computationally prohibitive for flexible interactive applications, due to the heavy-weight backbones associated with such models. 
To overcome this challenge, we introduce a novel approach that is known as semantic feature imitation processing. 
This process enables direct rendering of semantic-aware features similar to rendering colors in NeRF. 
By leveraging this process, the pretrained NeRF model can learn and incorporate meaningful semantic patterns from the output of the semantic model backbones, allowing it to accurately segment rendered images using lightweight decoders of the semantic models efficiently. 
For instance, with SAM~\cite{SAM}, which is the state-of-the-art image segmentation model, our framework could accelerate the segmentation process by 16 times with comparable mask quality. 
By leveraging this innovative approach, our framework unlocks tremendous performance potential for powering real-time interactive applications with ease.

Our proposed framework offers several key advantages. 
First, it eliminates the need for expensive segmentation backbones, resulting in a significant speed up of segmentation processing and facilitating human-scene interaction. 
Second, our semantic imitation module is pluggable and independent of the original NeRF module, thus preserving the rendering quality without compromise.
Finally, our approach is model-agnostic to both NeRF and perception models, enabling it to seamlessly integrate with advanced models in the future. 
Overall, our method presents a practical and flexible solution for enhancing NeRF-based real-world applicability and effectiveness.

\section{Related Work}

\subsection{NeRF for 3D representation}

Neural Radiance Fields (NeRF)~\cite{mildenhall2020nerf} has gained significant attention and led to rapid progress for photo-realistic novel view synthesis.
Various works are proposed to enhance different aspects of NeRF, like improving the quality of rendering image~\cite{barron2021mip,NeRF-SR,ZipNeRF}, decreasing training and inference time~\cite{SNeRF,mueller2022instant,MERF}, expanding the applicable range~\cite{barron2022mipnerf360,DNeRF,tang2022compressible}, and generalizing with few-shot settings~\cite{pixelNeRF,FreeNeRF}. 
NeRF has also been widely used in a variety of applications, including 3D-aware image generation~\cite{GIRAFFE,GIRAFFE_HD,tang2022real}, text-to-3D generation~\cite{poole2022dreamfusion,Magic3D}, and 3D shape generation~\cite{im2nerf,CodeNeRF}, pose estimation~\cite{bian2022nopenerf,BARF}. 
Our work aims to enhance semantic understanding ability of NeRF, especially on user interaction.

\subsection{2D Semantic Understanding}
Amazing progress has been made in image semantic understanding in the past few years. 
Detection Transformer (DETR)~\cite{DETR} and follow-ups~\cite{Cheng2021MaskedattentionMT,Jain2022OneFormerOT,Li2022MaskDT,chen2022d,chen2022group,chen2022conditional,meng2021conditional,chen2022groupv2,chen2022context,zhang2022cae} employ transformer-based architecture~\cite{Transformer} and have made significant advance in semantic understanding. 
Combined with large-scale vision-language pretraining models like CLIP~\cite{CLIP}, open-vocabulary segmentation methods~\cite{DBLP:conf/eccv/GhiasiGCL22,DBLP:conf/cvpr/HuynhKLGE22,Rao2021DenseCLIPLD,DBLP:conf/cvpr/XuMLBBKW22,wang2023visionllm} can perform segmentation from human language prompts. 
To facilitate user interactions, some works~\cite{DBLP:conf/cvpr/XuPCYH16,DBLP:journals/corr/abs-2210-11006,DBLP:conf/cvpr/ChenZZDQZ22} propose to use strokes or clicks as input for segmentation.
More recently, Segment Anything Model (SAM)~\cite{SAM} trains on large-scale datasets and achieves strong zero-shot performance given various visual prompts like clicks or boxes. 
X-Decoder~\cite{X-Decoder} proposes a unified approach to support various types of segmentation and vision-language tasks including open-vocabulary segmentation. 
SEEM~\cite{SEEM} further includes visual and audio prompts into a join visual-semantic space and enables composition of different types of prompts. 
However, these methods usually rely on heavy-weight segmentation backbones to extract semantic features from images, which slows down the inference speed.
We propose to bypass the segmentation backbones with our feature imitation module, thus accelerating various semantic tasks in 3D space.

\subsection{3D Semantic Understanding}
Compared to semantic understanding in 2D images, 3D semantic understanding is more complex. 
Existing methods~\cite{DBLP:conf/cvpr/HuangWN18,DBLP:conf/nips/YangWCHWMT19,DBLP:conf/cvpr/VuKLNY22,PLA,tang2022not,chen20203d,chen2020real,MinZhong2023MaskGroupHP,tang2022point} mainly focus on closed set segmentation where scenes are represented by point clouds or voxels. 
Recently, some works explore to apply NeRF in 3D semantic understanding, including object segmentation~\cite{NeRF-SOS,DBLP:conf/nips/LiuCYTT22,NeuralDiff,ONeRF}, panoptic segmentation~\cite{PanopticLifting,PanopticNeRF}, 3D semantic segmentation~\cite{DBLP:conf/iccv/ZhiLLD21}, part segmentation~\cite{SegNeRF}, text-based segmentation~\cite{DFF,lerf2023}, and interactive segmentation~\cite{N3F,ren-cvpr2022-nvos,ISRF}.
In particular, NVOS~\cite{ren-cvpr2022-nvos} segments objects in neural volumetric representations using positive and negative user strokes inputs. ISRF~\cite{ISRF} also uses user strokes as input and distills 2D semantic features from a large self-supervised pretrained model to NeRF and use nearest neighbor feature matching to segment objects. 
By fusing CLIP embedding into a NeRF, LERF~\cite{lerf2023} supports using human language to localize a wide variety of queries in 3D scenes. 
Our work is able to perform both click- and text-based 3D segmentation using different 2D perception backbones, while also being much faster compared to previous works and achieving real-time interaction.
\section{Preliminaries}

\subsection{NeRF}

The Neural Radiance Fields (NeRF) approach, introduced by Mildenhall et al.~\cite{mildenhall2020nerf}, employs a 5D function $f_\Theta$ to depict a 3D volumetric scene. 
This function takes a 3D coordinate $\mathbf{x} = (x,y,z)$ and a 2D viewing direction $\mathbf{d} = (\theta, \phi)$ as inputs, and outputs a volume density $\sigma$ with an emitted color $\mathbf{c} = (r, g, b)$.
For a ray $\mathbf{r}$ that starts at $\mathbf{o}$ and follows the direction $\mathbf{d}$, we sample points $\mathbf{x}_i = \mathbf{o} + t_i \mathbf{d}$ along the ray in sequence, and use $f_\Theta$ to retrieve densities ${\sigma_i}$ and colors ${\mathbf{c}_i}$.
The color of the pixel associated with the ray can then be estimated using numerical quadrature:

\begin{equation}
    \label{eq:rendering}
    \mathbf{\hat C}(\mathbf{r}) = \sum_i T_i \alpha_i \mathbf{c}_i, T_i = \prod_{j < i} (1 - \alpha_i), \alpha_i = 1 - \exp(-\sigma_i \delta_i), \delta_i = t_{i+1} - t_i
\end{equation}

Here, $\delta_i$ denotes the step size, $\alpha_i$ is the opacity, and $T_i$ is the transmittance.
The process of volume rendering is differentiable, which allows NeRF to be optimized using only 2D image supervision. This is done by minimizing the L2 difference between the predicted color of each pixel $\mathbf{\hat C}(\mathbf{r})$ and the actual color $\mathbf{C}(\mathbf{r})$ from the image:

\begin{align}
    \label{loss:ori}
    \mathcal{L}_{\text{NeRF}} = \sum_{\mathbf{r}} || \mathbf{C}(\mathbf{r}) - \mathbf{\hat C}(\mathbf{r}) ||_2^2
\end{align}

\subsection{Mesh Extraction from NeRF} 

The implicit volumetric representations of NeRF differ significantly from the widely-adopted polygonal meshes and lack support from common 3D software and hardware, making their rendering and manipulation inefficient. 
Recent works~\cite{tang2022nerf2mesh,bakedsdf,chen2022mobilenerf,munkberg2022extracting} explore using surface meshes for accelerated rendering.
For example, NVdiffrec~\cite{munkberg2022extracting} uses differentiable rendering to optimize a tetrahedron grid for geometry, while the texture can be encoded with in a multi-scale hashgrid~\cite{mueller2022instant}.
NeRF2Mesh~\cite{tang2022nerf2mesh} further extends mesh representation into unbounded scenes by optimizing a coarse mesh extracted from NeRF.
Our method is agnostic to the underlying 3D representation and can be used in such mesh-based setting too, which leads to faster RGB rendering and also enables mesh segmentation.

\subsection{Large Perception Models}

Recently, large scale vision datasets and transformer architecture have enabled training of strong perception models like SAM~\cite{SAM} and X-Decoder~\cite{X-Decoder} for 2D dense semantic understanding.
These models typically consist of an image backbone and a prompt decoder.
The image backbone $\text{Enc}$ usually adopts a heavy-weight transformer ($e.g.,$ Vit-Huge~\cite{vit}) to encode semantic features from input image $\mathbf{I}$, while the prompt decoder $\text{Dec}$ embeds different input prompts $p$ such as point clicks or text descriptions to predict semantic masks $\mathbf{M}$:
\begin{equation}
    \mathbf{M} = \text{Dec}(\text{Enc}(\mathbf{I}), p).
\end{equation}
\section{Method}

In this section, we introduce our pipeline to perform interactive 3D segmentation.
First, we propose a semantic feature imitation module to replace heavy-weight segmentation backbones by directly rendering semantic features in Section~\ref{sec:sfi}.
Next, we explore different loss functions for single-scale and multi-scale feature imitation in Section~\ref{sec:loss}.
Then, we discuss several key designs including camera augmentation and caching to enhance feature imitation quality in Section~\ref{sec:details}.
Finally, we show how to perform user interaction in our graphic user interface and potential applications in Section~\ref{sec:gui}.

\begin{figure}[t]
    \centering
    \includegraphics[width=\linewidth]{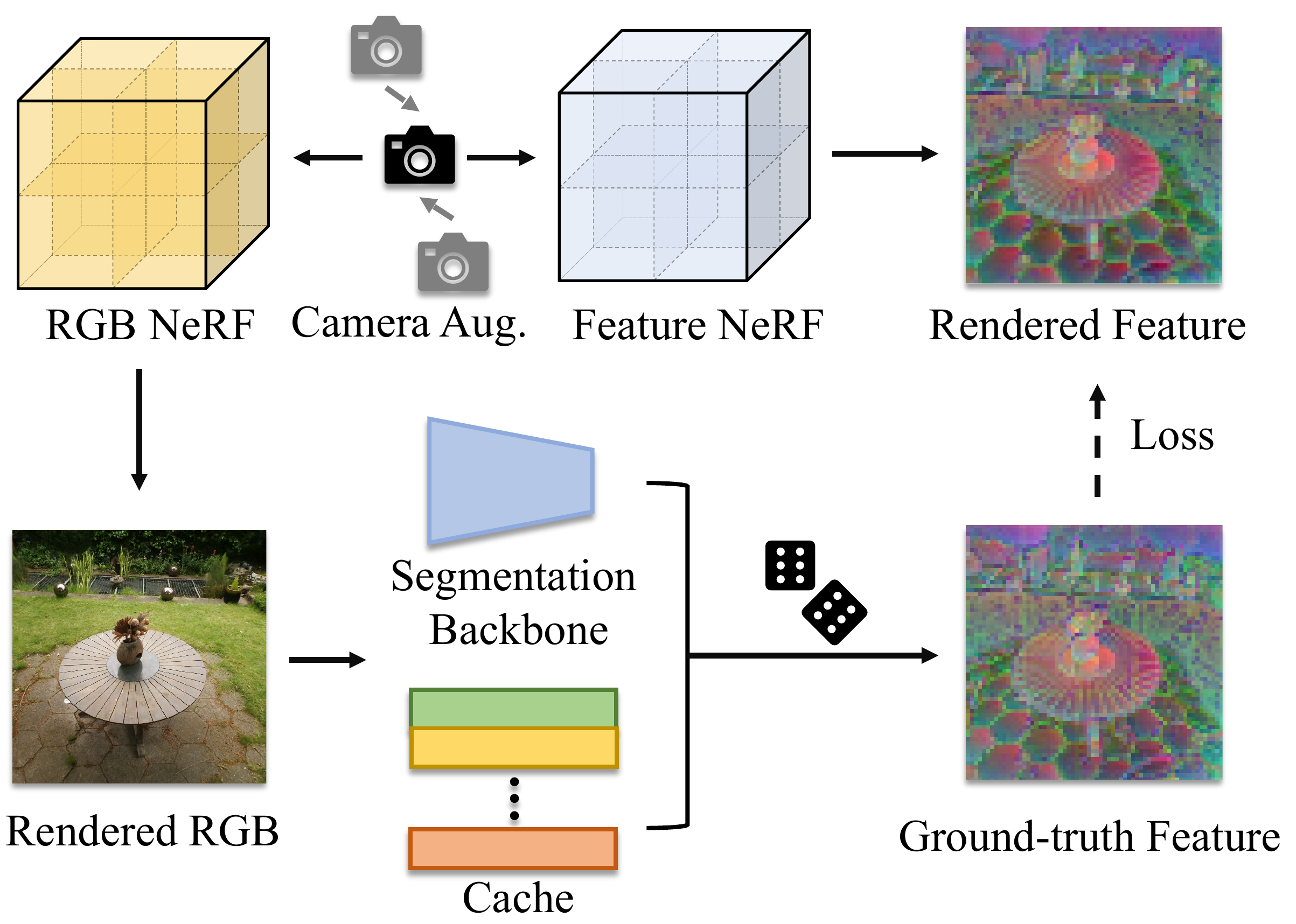}
    \caption{
    \textbf{Semantic feature imitation training.} 
    We visualize the high-dimensional semantic features by rendering the first three channels as RGB.}
    \label{fig:arch}
\vspace{-3mm}
\end{figure}

\subsection{Semantic Feature Imitation}
\label{sec:sfi}

We propose to render the semantic features from neural radiance fields directly, eliminating the need for the forward process of the segmentation backbone.
The overall training procedure is shown in Figure~\ref{fig:arch}.
Our method is model-agnostic and could be applied to a broad range of NeRF. In implementation, we choose grid-based NeRF~\cite{mueller2022instant,tang2022nerf2mesh} to enhance efficiency.
Assume we have a trained NeRF model that predicts the density $\sigma$ and color $\mathbf{c}$ at each 3D location $\mathbf{x}$:
\begin{align}
    \sigma = \Phi(\text{MLP}(E^\text{geo}(\mathbf{x}))), \\ 
    \mathbf{c} = \Psi(\text{MLP}(E^\text{rgb}(\mathbf{x}))), 
\end{align}
where $\Phi$ is the exponential activation~\cite{mueller2022instant} that promotes sharper surface and $\Psi$ refers to the sigmoid activation. 
$E^\text{geo}$ and $E^\text{rgb}$ are learnable feature grids to represent the 3D field.

Our semantic feature imitation module is totally pluggable to the original RGB and density field in NeRF, which are fixed during semantic feature imitation training.
Given a camera view, we first render the RGB image $\mathbf{I}$ with NeRF.
We then use the perception model's backbone to extract semantic features $\mathbf{F} = \text{Enc}(\mathbf{I}) \in \mathbb{R}^{C \times h \times w}$, where $C$ is the feature dimension and $h, w$ are the feature height and width (usually smaller than image resolution $H, W$). 
These 2D feature maps can be used to supervise 3D feature grid just like the RGB information.
We reuse the density information and render the semantic feature along each ray $\mathbf{r}$ using numerical quadrature:
\begin{equation}
\label{eq:render_feature}
    \mathbf{\hat F}(\mathbf{r}) = \text{MLP}(\sum_i T_i \alpha_i E^\text{sem}(\mathbf{x}_i)) \\
\end{equation}
where $\mathbf{\hat F}(\mathbf{r})$ is the imitated feature and $E^\text{sem}$ is the learnable semantic feature grid.
As the feature channels $C$ are usually much higher than 3-channel RGB, we move the non-linear MLP after performing quadrature following~\cite{MERF,hedman2021snerg}.
Also, we perform feature rendering directly at the smaller feature resolution $h \times w$ after RGB rendering, which further saves the computation. 
In cases when we can extract a mesh surface from the NeRF~\cite{tang2022nerf2mesh}, we only need to sample one surface point $\mathbf{x}_s$ per ray and simplify Equation~\ref{eq:render_feature} to:
\begin{equation}
\label{eq:render_feature_mesh}
    \mathbf{\hat F}(\mathbf{r}) = \text{MLP}(E^\text{sem}(\mathbf{x}_s)) \\
\end{equation}

Different perception models may require different number of features maps for the decoder.
For single-scale decoder as used in SAM~\cite{SAM}, we only need to imitate one feature map $\mathbf{F}$. 
Other models like X-Decoder~\cite{X-Decoder} use multi-scale feature maps $\{\mathbf{F}_i\}, i \in [0, 3]$.
In such cases, we share the feature grid $E^\text{sem}$ and use different heads $\text{MLP}_{i}, i \in [0, 3]$ to predict multi-scale features, which helps to exploit the cross-scale correlation and reduce total parameters.

During inference, we render all the necessary semantic feature maps $\mathbf{\hat F}$ of the test image, and apply the perception model's decoder to generate mask predictions $\mathbf{\hat M} = \text{Dec}(\mathbf{\hat F})$.

\subsection{Loss Function}
\label{sec:loss}

\noindent \textbf{Single-Scale Decoder.}
For single-scale decoder models, we directly minimize the MSE loss between the rendered semantic features $\mathbf{\hat F}$ and the ground-truth semantic features $\mathbf{F}$:
\begin{equation}
    \mathcal{L}_{\text{single}} = \text{MSE}(\mathbf{\hat F}, \mathbf{F}) = \frac 1 N \sum_{\mathbf{r}} || \mathbf{F}(\mathbf{r}) - \mathbf{\hat F}(\mathbf{r}) ||_2^2
\end{equation}
where $N$ is the number of rays per training step.

\vspace{1mm}
\noindent \textbf{Multi-Scale Decoder.}
For multi-scale decoder models, the feature maps at different scales are usually correlated to represent the same underlying image~\cite{long2015fully,lin2017feature}.
Based on this observation, we introduce an additional loss term called the cross-scale correlation loss. This loss term promotes to learn the correlation between semantic features at different scales and aids in capturing contextual information. 
We first calculate the cosine similarity map between the $i$-th and the $j$-th multi-scale feature maps and obtain a similarity maps $\mathbf{S}_{ij} \in \mathbb{R}^{C \times s_i \times s_j}$, where $s_i, s_j$ represent the number of pixels in the two feature maps. This cosine similarity map can then be used as an extra supervision:
\begin{align}
    \mathcal{L}_{\text{multi}} &= \mathcal{L}_{\text{single}} + \mathcal{L}_{\text{cross}} \\ 
    \mathcal{L}_{\text{single}} &= \sum_{i}\text{MSE}(\mathbf{\hat F}_i, \mathbf{F}_i) \\
    \mathcal{L}_{\text{cross}} &= \sum_{i}\sum_{j>i} \text{MSE}(\mathbf{\hat S}_{ij}, \mathbf{S}_{ij})
\end{align}
$\mathbf{\hat S}_{ij}$ is the similarity map of predicted feature maps.
In the experiment, we find this cross-scale correlation loss could improve feature imitation quality and accelerate convergence.

\subsection{Training Details}
\label{sec:details}

\noindent \textbf{Camera Augmentation.}
Since we are able to synthesize RGB images from arbitrary camera pose with the pretrained NeRF, we augment the training dataset by interpolating between the original training camera poses for feature imitation training.
This technique is widely used to distill NeRF~\cite{cao2022real,wang2022r2l}, which shares a similar pipeline with ours.
By capturing a more diverse range of viewpoints, we found that camera augmentation helps to enhance the robustness of our semantic feature imitation and results in more smooth segmentation predictions. 
During training, we perform on-the-fly rendering of the novel RGB images and then extract the corresponding semantic features by the preception model as supervision.

\vspace{1mm}
\noindent \textbf{Caching Mechanism.} 
The above on-the-fly feature extraction training leads to heavy burden as we run the heavy perception model per training step.
To mitigate the large computational cost associated with the segmentation backbone, we employ a caching mechanism. 
During training, we maintain a cache that stores the camera view, RGB image, and the corresponding feature map.
For the initial steps, we forward novel view images to the segmentation backbone and cache the resulting feature maps. 
In the following steps, we adopt a randomized approach to determine whether to use a cached feature map or sample a new feature map as the supervision. 
By randomly choosing between the cached feature map and the online-generated feature map, we strike a balance between reusing cached information and incorporating fresh data during training.
To maintain efficiency and manage cache memory, we make the cache obey the first-in-first-out (FIFO) rule, where the oldest entries in the cache are replaced with new ones when the cache reaches its capacity.
By employing this caching mechanism, we significantly reduce the computational cost and accelerate the training, while maintaining similar performance.

\begin{figure}[t]
    \centering
    \includegraphics[width=\linewidth]{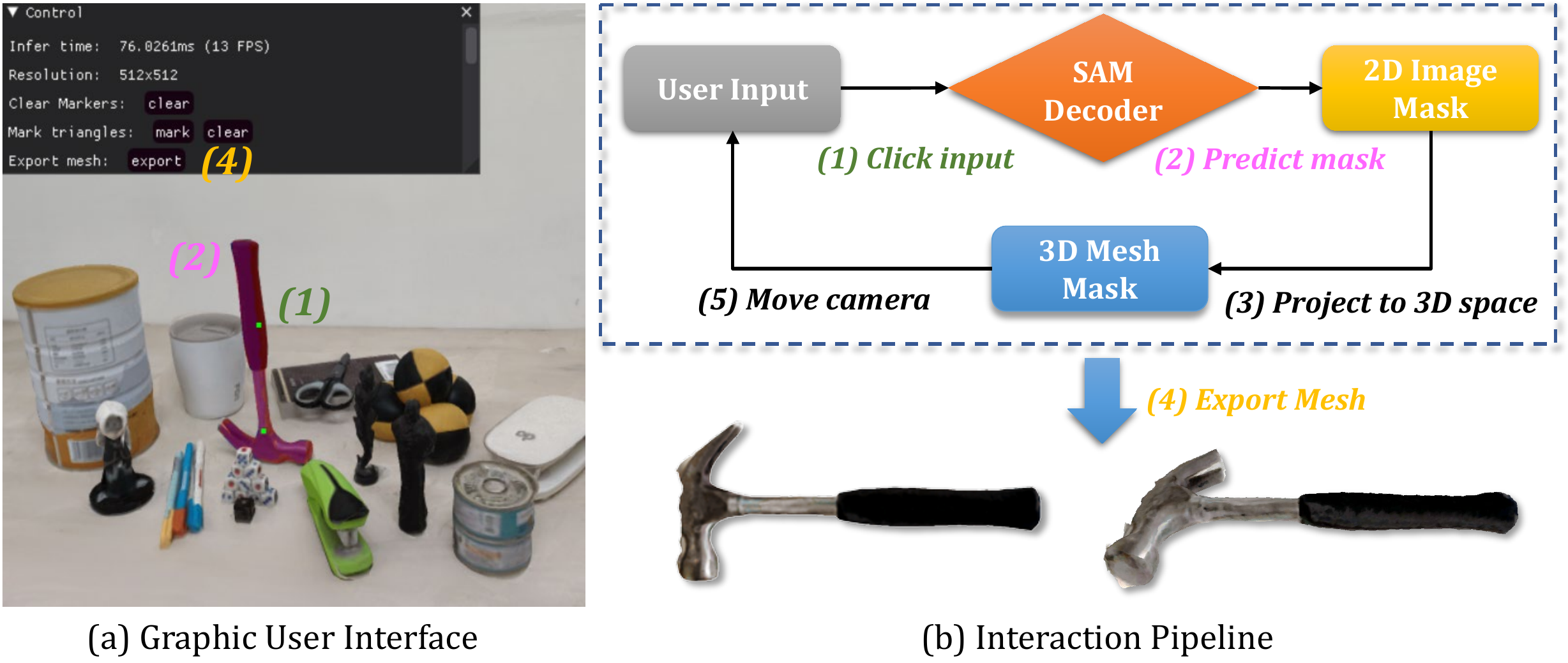}
    \caption{
    \textbf{GUI and interaction pipeline.} We design a GUI that allows user interaction in real-time to perform 2D image segmentation and 3D mesh segmentation.
    }
    \label{fig:gui}
\end{figure}

\begin{figure*}[t!]
    \centering
    \includegraphics[width=0.97\linewidth]{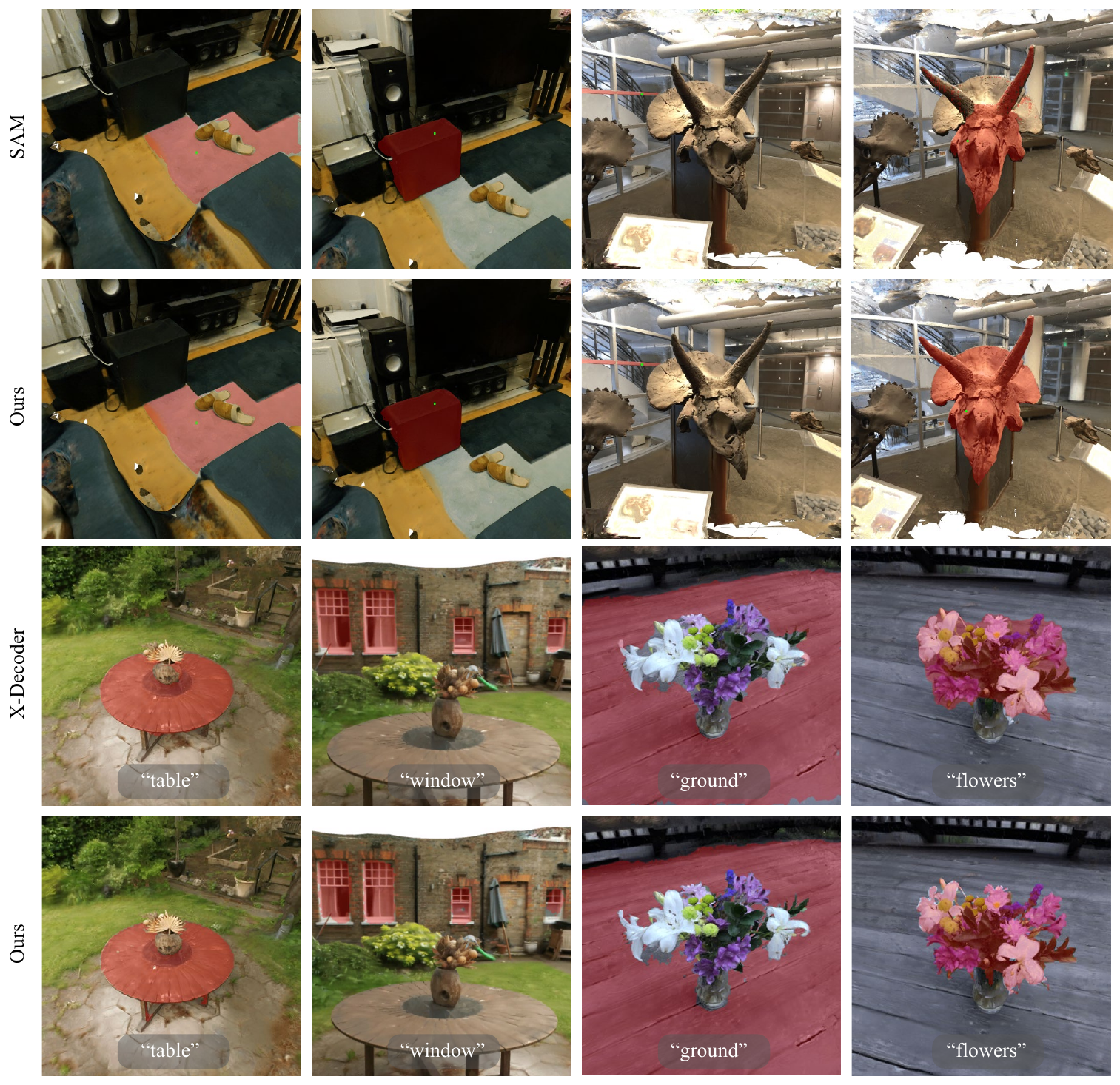}
    \caption{
    \textbf{Semantic segmentation results}.
    The click input is shown in {\color{green} green} dot and the segmentation mask in {\color{red} red}. The text input is shown in {\color{gray} gray box}.
    Our method achieves comparable segmentation quality compared to the perception models. We observe in some cases, our results achieve more robust and desirable mask benefits from the 3D consistency.
    }
    \label{fig:seg}
\vspace{-2mm}
\end{figure*}

\subsection{User Interactivity}
\label{sec:gui}

To demonstrate the effectiveness of our method, we implement a GUI for user interaction as shown in Figure~\ref{fig:gui}.
The GUI allows users to drag and view the 3D scene in real-time.
For any camera view, the user can input a prompt (\textit{e.g.}, click a point on the viewport), and our model renders the 2D segmentation mask overlayed on the RGB image.
Then the user can choose to project the 2D mask onto 3D mesh surfaces, so the mask can be rendered from other camera views.
For click points, we also project it to 3D so it can be automatically tracked when changing the camera view.
By repeating the above process from several camera views, it's easy to get the targeted object fully segmented from the 3D space.
Finally, the user can export the segmented mesh and also the texture maps.
\section{Experiments}

\subsection{Implementation Details}

\noindent \textbf{Training Setting.}
We use NeRF2Mesh~\cite{tang2022nerf2mesh} as the framework for our NeRF training and Mesh extraction.
The training of NeRF takes $10,000$ steps, with each step containing approximately $2^{18}$ points. 
An exponentially decayed learning rate schedule ranging from $0.01$ to $0.001$ is employed. 
A coarse mesh is then extracted and finetuned for additional $5,000$ steps.
Then, the training of feature grid takes $10,000$ steps.
At each step, we have $75\%$ chance to use a cached camera view, and $25\%$ chance to sample new camera view. 
The cache size is set to $256$.
We use the Adam~\cite{kingma2014adam} optimizer in all stages. 
All experiments are conducted on a single NVIDIA V100 GPU.

\noindent \textbf{Datasets.}
We choose the widely used Mip-NeRF 360~\cite{barron2022mipnerf360} dataset and LLFF~\cite{mildenhall2019llff} dataset to evaluate our method.
Mip-NeRF 360~\cite{barron2022mipnerf360} dataset contains 4 indoor scenes and 3 outdoor scenes captured in 360 degree.
LLFF~\cite{mildenhall2019llff} dataset contains 8 scenes captured in a forward-facing manner.
Furthermore, we use some self-captured custom data to demonstrate the generalization ability of our method.

\subsection{Segmentation Efficiency}
We first analyse the segmentation efficiency in Table~\ref{tab:efficiency}. Compared to SAM~\cite{SAM}, our imitation model delivers a 52$\times$ (624 ms $\rightarrow$ 12 ms) increase in feature encoding speed, leading to a 16$\times$ (1.53 FPS $\rightarrow$ 24.39 FPS) boost in overall rendering speed. This makes our method the first real-time 3D click-based segmentation model capable of facilitating smooth user interaction at a $512 \times 512$ resolution with a contemporary NVIDIA GPU.
In prompt-based segmentation with X-decoder~\cite{X-Decoder}, our model still manages to double the feature encoding speed, despite that X-Decoder's image backbone is relatively small. However, the oversized decoder of X-Decoder has become a bottleneck, limiting the FPS to around 5.
Still, our method is considerably faster compared to other prompt-based models like LeRF~\cite{lerf2023} which runs around 1 FPS.
Our successful validation with prompted-based X-Decoder suggests that our method has the potential to perform well given the availability of more suitable prompt-based segmentation models. An ideal model might feature a larger backbone and a more lighweight decoder.

\subsection{Qualitative Evaluation}
We demonstrate the feature imitation quality of our method through performing segmentation in challenging realistic 3D scenes. In Figure~\ref{fig:seg}, we present the results of click- and text-based segmentation. Our imitation model achieves a segmentation quality comparable to that generated by pretrained segmentation models, even in intricate scenarios such as foliage and slender areas. Our model also demonstrates proficiency in segmenting different regions based on language prompts. Notably, we find our results tend to be more robust in capturing the entirety of the object than X-Decoder, due to the inherent 3D consistency of the features.

\begin{figure}[ht]
    \centering
    \includegraphics[width=\linewidth]{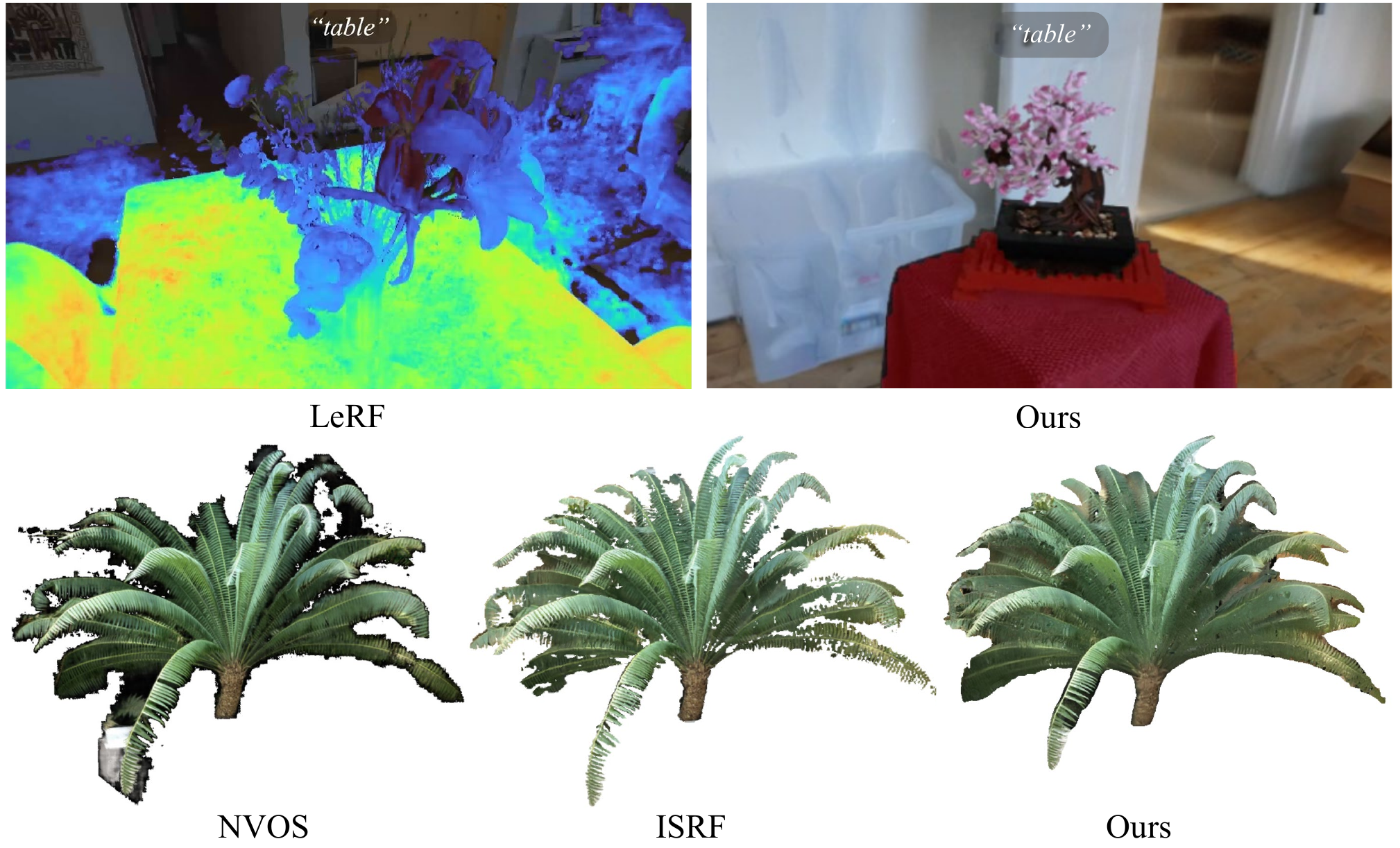}
    \caption{
    \textbf{Comparison with other methods.} We compare against other methods using prompts or strokes as input for 3D semantic understanding. Since the data used by LeRF~\cite{lerf2023} is not publicly available, we choose a similar scene from the Mip-NeRF 360 dataset~\cite{barron2022mipnerf360}.
    }
    \label{fig:comparison}
\end{figure}

We also compare our approach with recent methods~\cite{lerf2023,ISRF,ren-cvpr2022-nvos} in Figure~\ref{fig:comparison}. Unlike LeRF~\cite{lerf2023}, our method is capable of generating masks with distinct boundaries. Compared with ISRF~\cite{ISRF} and NVOS~\cite{ren-cvpr2022-nvos} that utilize strokes as input, our method achieves better or comparable quality only using one point click as input.

\begin{table}[t]
\begin{center}
\renewcommand{\arraystretch}{1.1}
\begin{tabular}{l|c|c|c}
\toprule

                           &    Method      & SAM & X-Decoder \\
\hline
RGB Rendering (ms)              & -        & 17  & 17 \\ 
\hline                           
\multirow{2}{*}{Feature Encoding (ms)}  & Original & 624  &  73  \\
                           & Ours     & \textbf{12}  &  \textbf{38}  \\
\hline                           
Feature Decoding (ms)           & -        & 8   & 155 \\ 
\hline
\multirow{2}{*}{FPS}       & Original & 1.53  &  4.07  \\
                           & Ours     & \textbf{24.39} ({\color{blue} 16$\times$})  & \textbf{4.71} ({\color{blue} 1.2$\times$}) \\

\bottomrule
\end{tabular}

\end{center}
\caption{
\textbf{Inference efficiency}.
We report the inference time of three major steps in milliseconds (ms), and overall FPS.
}
\vspace{-0.5cm}
\label{tab:efficiency}
\end{table}

\begin{figure}[t]
    \centering
    \includegraphics[width=\linewidth]{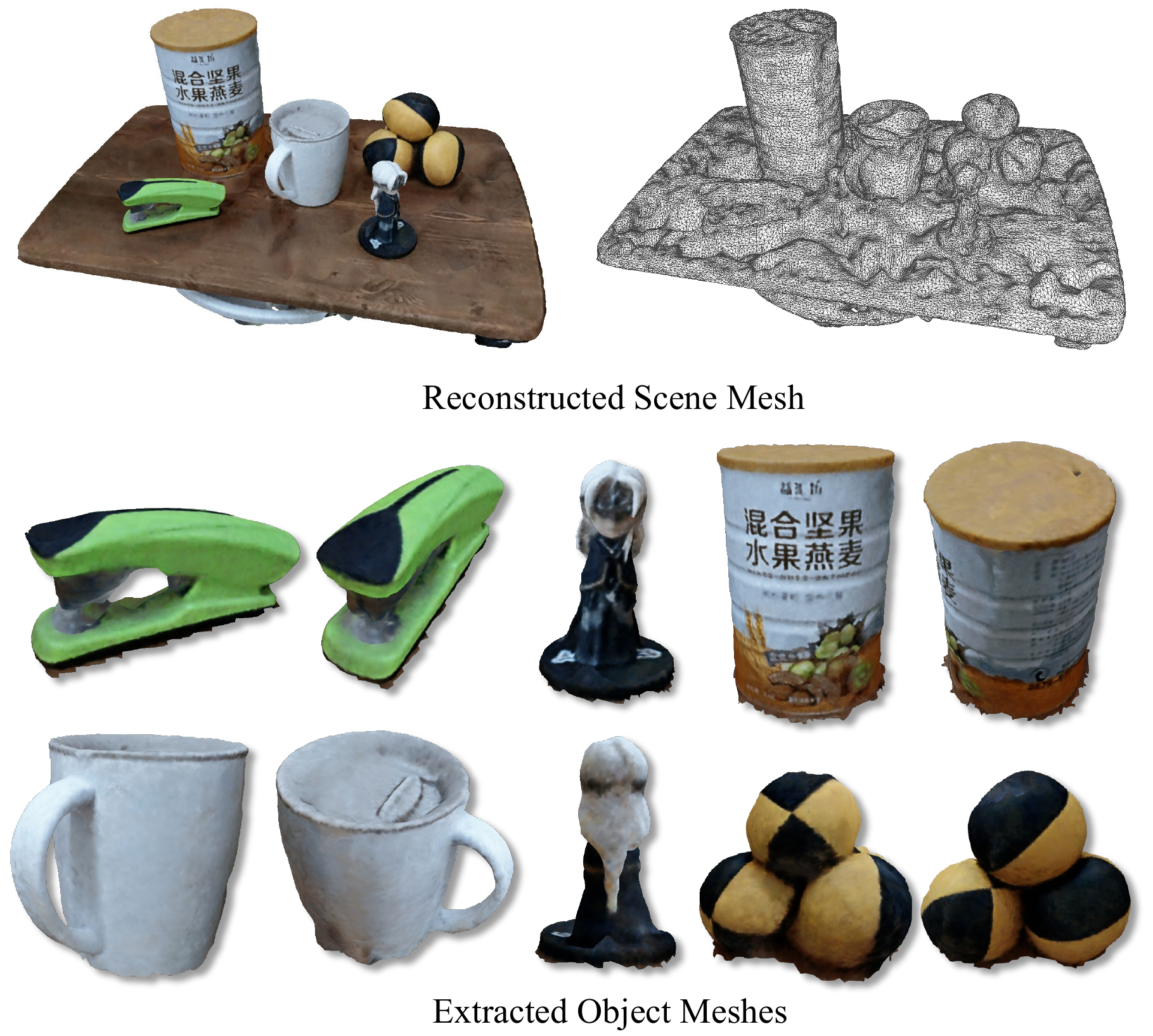}
    \caption{
    \textbf{Mesh segmentation results}.
    Our method allows mesh segmentation by projecting the 2D masks to 3D surfaces. With simple interactions in our GUI, users can get ready-to-use single object meshes easily.
    }
    \label{fig:meshing}
\vspace{-2mm}
\end{figure}

\subsection{Quantitative Evaluation}
\label{sec:quantitative}
To further evaluate the quality of feature imitation beyond qualitative analysis, we conduct a quantitative evaluation.
We report the Intersection over Union (IoU) between the predicted mask and the target mask (produced by the pretrained segmentation model). 
For SAM, we uniformly sample $5 \times 5$ points in the image, and each point corresponds to a segmentation result.
For X-Decoder, we design several prompts for each scene based on the objects present, and each prompt corresponds to a segmentation result. For example, the test prompts for the scene ``garden'' in Mip-NeRF 360 dataset~\cite{barron2022mipnerf360} are ``table'', ``window'' and ``grass''. 
Our imitation model achieves 82.2\% and 74.9\% IoU for SAM and X-Decoder, respectively.

\begin{table*}
  \centering
    \renewcommand{\arraystretch}{1.1}
  \begin{tabular}{ccccccc}
  \toprule
      & Camera Aug. & Correlation & Caching & Feature Loss $\downarrow$ & Mask IoU $\uparrow$ & Training Time $\downarrow$ \\
    \hline
    \multirow{3}{*}{SAM} 
      & \checkmark & - & \checkmark & 0.0026 & 0.822 & 90 min   \\
      & - & - & \checkmark & 0.0039 & 0.746 & 89  min \\
      & \checkmark & - & - & 0.0025 & 0.824 & 303 min \\
      \hline
    \multirow{4}{*}{X-Decoder} 
      & \checkmark & \checkmark & \checkmark & 0.744 & 0.749 & 30 min   \\
      & - & \checkmark & \checkmark & 0.751 & 0.742 & 30 min \\
      &  \checkmark & - & \checkmark & 0.759 & 0.741 & 28 min \\
      & \checkmark & \checkmark & - & 0.743 & 0.751 & 42 min \\
    \bottomrule
  \end{tabular}
  \caption{\textbf{Ablation studies.} We report feature MSE loss value, mask IoU, and training time to compare different settings. Please note that the feature value ranges of SAM and X-Decoder are different, so the loss of these two methods cannot be directly compared.}
  \label{tab:ablation}
  \vspace{-4mm}
\end{table*}




\subsection{Mesh Segmentation}
We showcase the results of mesh segmentation in Figure~\ref{fig:meshing}. In a complex scene composed of multiple objects, we project the 2D segmentation masks onto 3D surfaces from various viewpoints to achieve mesh segmentation. This approach allows us to extract single object meshes with simple, controllable user interaction. 
Please check our demonstration video for a practical usage example.
Since our pipeline's ultimate output representation is a mesh, we pave the way for a variety of downstream applications based on the extracted single-object meshes. For instance, we can carry out texture editing and model compositions using common 3D software tools, as demonstrated in Figure~\ref{fig:teaser}. 

\subsection{Ablations}
We perform ablation studies to verify the designs of our method. We report two metrics on the test set: (1) MSE loss between the imitated feature and the ground-truth feature from the segmentation backbone. (2) Mask IoU that introduced in section~\ref{sec:quantitative}.

\noindent \textbf{Camera Augmentation.} 
We generate novel views through interpolation, thereby expanding the dataset during training. 
It can be observed that camera augmentation are helpful in reducing feature loss and improving final mask IoU for both SAM and X-decoder backend, demonstrating its effectiveness. 

\noindent \textbf{Caching.} 
We could reduce the cost of running segmentation backbone by reutilizing cached ground-truth features, which is the major training time bottleneck.
In Table~\ref{tab:ablation}, we find that our caching mechanism does not markedly affect the feature imitation quality, while significantly accelerating training.
This enhancement is particularly conspicuous in models with a heavy backbone such as SAM~\cite{SAM}, where we can decrease the training time by a factor of 3.

\noindent \textbf{Cross-scale Correlation Loss.}
We also verify the effectiveness of the proposed cross-scale correlation loss. 
By constraining the correlations between multi-scale feature maps, we observe smaller feature loss for X-Decoder~\cite{X-Decoder} and also improved IoU.

\section{Limitations}

\begin{figure}[h]
    \centering
    \includegraphics[width=\linewidth]{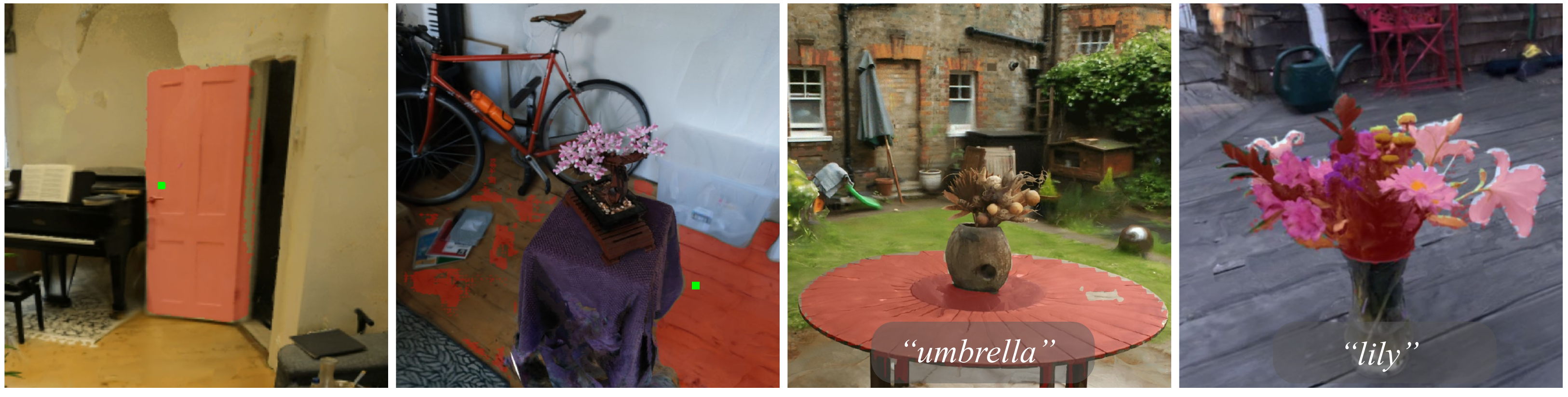}
    \caption{
    \textbf{Failure cases.} We show some cases when our method cannot produce satisfactory segmentation masks for both click- and text-based prompts.
    }
    \label{fig:failure}
\end{figure}

In Figure~\ref{fig:failure}, we show some failure cases of our method.
For click-based segmentation, the mask could be imperfect and produce multiple unconnected regions. 
These are also the problems with the original SAM and can be solved by using multiple positive points and also negative points, or other input form like bounding boxes~\cite{SAM}.
For prompt-based segmentation, our performance is dependent on the original model's capability. We found that X-Decoder~\cite{X-Decoder} could fail to correctly recognize concept like ''umbrella`` but select the round table, or distinguish between close concepts like ''lily`` or ''flowers``. 
For mesh segmentation, we also rely on the NeRF backbone for mesh extraction.
These problems may be solved using more powerful perception models and NeRF backbones.

\section{Conclusions}
In this paper, we introduce a novel feature imitation pipeline designed to enhance NeRF with 2D perception models and accomplish 3D perception tasks. By substituting the heavy backbone of SAM with a feature rendering module, our model operates at 16$\times$ the speed of its counterparts while maintaining comparable quality. We devise several techniques to enhance imitation quality, including camera augmentation and cached training. Our pipeline has been validated using two state-of-the-art 2D models for click- and text-based segmentation, respectively. In addition, we have developed a graphical user interface to allow user interaction, which demonstrates the practical applicability of our method.
\bibliographystyle{ACM-Reference-Format}
\bibliography{main}

\end{document}